\documentclass[a4paper,conference]{IEEEtran}
\ifCLASSINFOpdf
\else
\fi
\usepackage[latin9]{inputenc}
\usepackage{amsmath, amssymb, bbm}
\usepackage{dsfont}
\usepackage{graphicx}
\usepackage{multirow}
\usepackage{pifont}
\usepackage{color,soul}
\usepackage{subcaption}
\usepackage{float}

\def\ie{\emph{i.e.}}

\def\etal{\emph{et al}}

\newcommand{\cmark}{\ding{51}}%
\newcommand{\xmark}{\ding{55}}%

\begin{document}
\newcommand{\func}[1]{\mathsf{#1}} 
\def \hh {\textbf{h}}
\def \bb {\textbf{b}}
\def \bg {\bb_g}
\def \bi {\bb_i}
\def \bj {\bb_j}
\def \simm {\left<\bg,\bi\right>}
\def \simmj {\left<\bg,\bj\right>}
\def \lr {\theta}
\def \bin {d}
\def \x {\mathbf{x}}
\def \ie {\textit{i.e.\,}}
\def \ptp {P_{\mathsf{tp}}}
\def \pfp {P_{\mathsf{fp}}}
%
\title{AggNet: Learning to Aggregate Faces for Group Membership Verification}

\author{
\IEEEauthorblockN{Marzieh Gheisari}
\IEEEauthorblockA{Inria, France
}
\and
\IEEEauthorblockN{Javad Amirian}
\IEEEauthorblockA{Inria, France
}
\and
\IEEEauthorblockN{Teddy Furon}
\IEEEauthorblockA{Univ Rennes, Inria, CNRS, IRISA\\
	France
}
\and
\IEEEauthorblockN{Laurent Amsaleg}
\IEEEauthorblockA{Univ Rennes, Inria, CNRS, IRISA\\France}
}


%


\maketitle

\begin{abstract}
In some face recognition applications, we are interested to verify whether an individual is a member of a group, without revealing their identity. Some existing methods, propose a mechanism for quantizing precomputed face descriptors into discrete embeddings and aggregating them into one group representation. However, this mechanism is only optimized for a given closed set of individuals and needs to learn the group representations from scratch every time the groups are changed. In this paper, we propose a deep architecture that jointly learns face descriptors and the aggregation mechanism for better end-to-end performances. The system can be applied to new groups with individuals never seen before and the scheme easily manages new memberships or membership endings. 
We show through experiments on multiple large-scale wild face datasets, that the proposed method leads to higher verification performance compared to other baselines.
\end{abstract}
\IEEEpeerreviewmaketitle


\section{Introduction}
\label{sec:intro}
Group membership verification assesses whether an individual is a member of a group for granting or refusing access to sensitive resources. This can be implemented through a two-phase process where identification is first performed, revealing the identity of the individual under scrutiny, followed by the verification that the identified individual is indeed a member of the claimed group.
That implementation breaks privacy: there is no fundamental reason to identify the individual before running the verification step.
It is fundamental to distinguish the members of the group from the non-members, but it does not require distinguishing members from one another.

Group membership verification first acquire templates of individuals (in this paper, face images) and then enroll them into a data structure stored in a server.
At verification time, that data structure is queried by a client with a new template, and the access is granted or refused.
Security assesses that the data structure is adequately protected so that an honest but curious server cannot reconstruct the individual templates.
Privacy requires that verification proceeds without disclosing the identity. 

A client acquires a fresh template and queries the server. Clients are trusted, but the server is honest but curious: It tries to reconstruct the enrolled templates or spy on the queries.
The design intends to prevent the server from reconstructing the individual template from the data structure while correctly determining whether or not a user is a member of the claimed group.


Gheisari \etal~\cite{Gheisari2019icassp} proposed a privacy-preserving group membership verification protocol quantizing identities's biometric templates into discrete embeddings and aggregating multiple embeddings into a group representation. That scheme has several desirable properties: it is cheap to run; quantization and aggregation fully succeed to make inversion and reconstruction so difficult that it impedes identification. It exhibits a trade-off between the security level and the group verification error rates.
That work, however, is deterministic in the sense that it uses precomputed face descriptors and  also follows a set of hard-coded rules as to how templates are quantized and aggregated and there is no learning. In another work of the same authors~\cite{Gheisari_2019_CVPR_Workshops}, the group representation is optimized for a given closed set of individuals.
In other words, the training and testing sets are the same group of people.
It means that any time a change in the group happens (a new member or a member leaving the group), the data structure has to be learned again from all the individual templates.
\medskip

\begin{figure*}[!htp]
	\centering
	\includegraphics[width=0.85\linewidth]{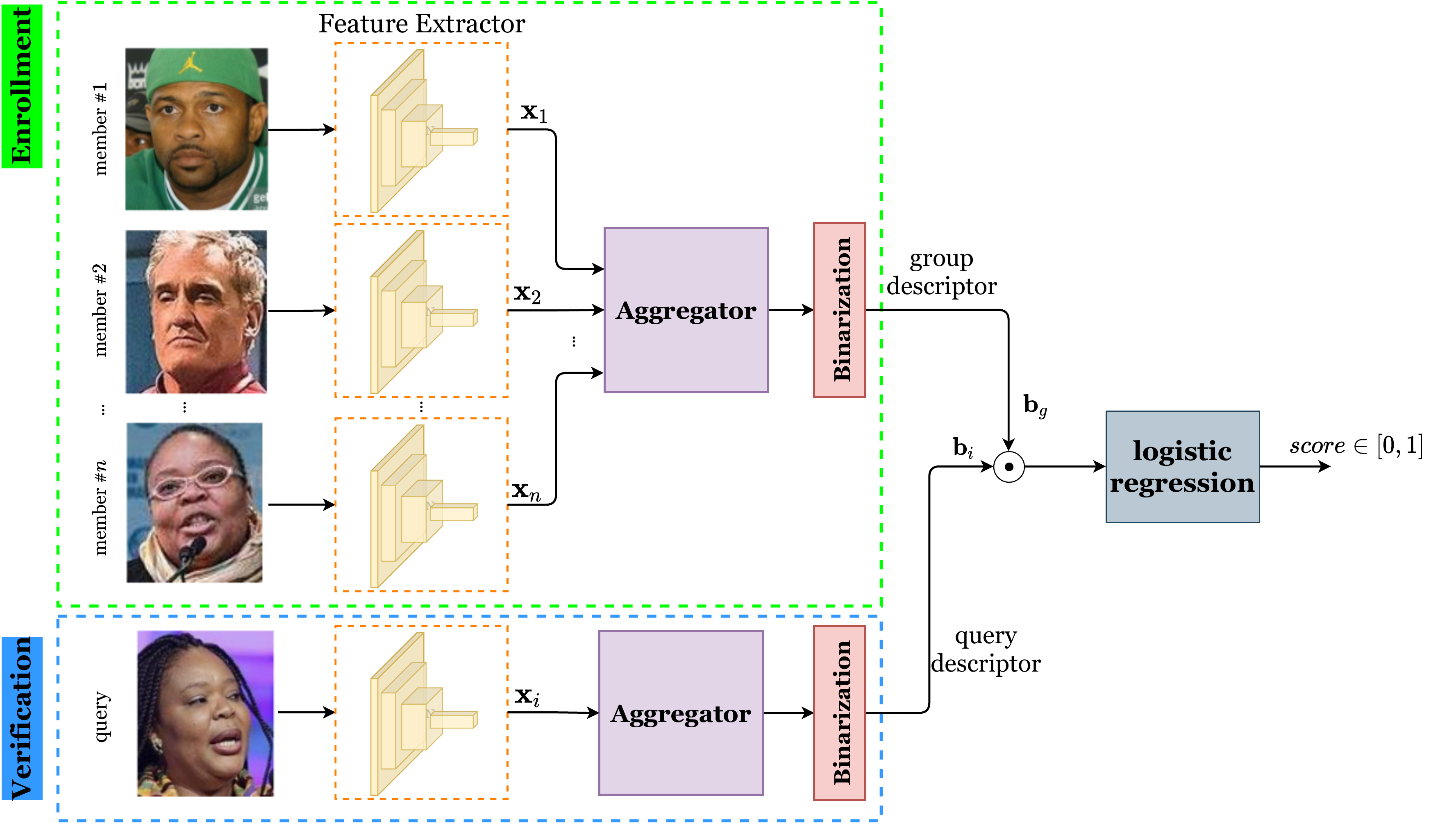}
	\caption{Overview of AggNet. At enrollment, the network is fed with faces of $n$ individuals to produce the group descriptor. At verification, the query descriptor is generated and the score computed by the logistic regression determines if the query belongs to this group.}
	\label{fig:aggnet-overview}
\end{figure*}

\noindent
The contributions of this paper are twofold:
\begin{itemize}
\item The proposed scheme jointly learns the face descriptors and the aggregation mechanism for better end-to-end performances. 
We consider an alternative objective function for training the classifier to maximize the AUC directly. We demonstrate that cross-entropy is not the most appropriate objective function for training a classifier when it is important to optimize the classifier's discriminative abilities over a range of thresholds. The system overview is shown in Fig.~\ref{fig:aggnet-overview}.

\item Once learned, these functionalities are sealed, but they can be applied to new groups with individuals never seen before, \ie not belonging to the training set.
This scheme easily manages new memberships or membership endings.

\end{itemize}

\section{Related Work}
\label{sec:related}
The concept of anonymous authentication for members of a group exists for a long time in cryptography~\cite{Schechter:1999qy}.  
As for biometric applications, our scenario is different from authentication, identification, and secret binding.
In those applications, security is provided at the server and/or client sides, but ultimately the user's identity is revealed.

The aggregation of signals into one representation is a common mechanism in computer vision. Approaches like BoW (Bag of Words)~\cite{Sivic:2003qp}, VLAD~\cite{jegou:inria-00633013} aggregate some local descriptors extracted from one image into a global description.  These popular encodings are also redesigned on top of convolutional networks. \cite{mohedano2016bags} proposed utilizing the BoW method to encode the convolutional features of CNNs. Arandjelovic \etal~\cite{arandjelovic2016netvlad} implemented VLAD with a learnable pooling layer, which is coined as NetVLAD. Soft assignment to multiple clusters is also proposed to train the pooling layer~\cite{radenovic2018fine}.

Zhong \etal~\cite{Zhong:2018aa} computes a single real-valued descriptor for the faces of celebrities that appear in the same picture. The query is a small set of face descriptors (usually two or three faces), and their system returns photos where this particular set of celebrities appear together.
Their performance is acceptable only when there are two faces per image, but degrades significantly when there are more faces.
We draw inspiration from these works in our work. However, we are dealing here with a  different problem since the query is a single face in our case.
On the other hand, our groups typically consist of more than two faces, and each face has possibly been captured under a different condition.

The aggregation of templates into one group representation is also well known in biometrics. 
For instance, in~\cite{Zhong:2018ab}, multiple faces captured from the same person are combined to gain robustness against poses, expression, and quality variations.
In our case, however, the unique faces of different individuals in the group need to be aggregated.
Last but not least, these approaches are designed to facilitate the retrieval of visually similar elements ; however, they do not offer any security or privacy.

\section{Our proposal: AggNet}
\label{sec:proposed}
\begin{figure*}
	\centering
	\includegraphics[width=0.9\linewidth]{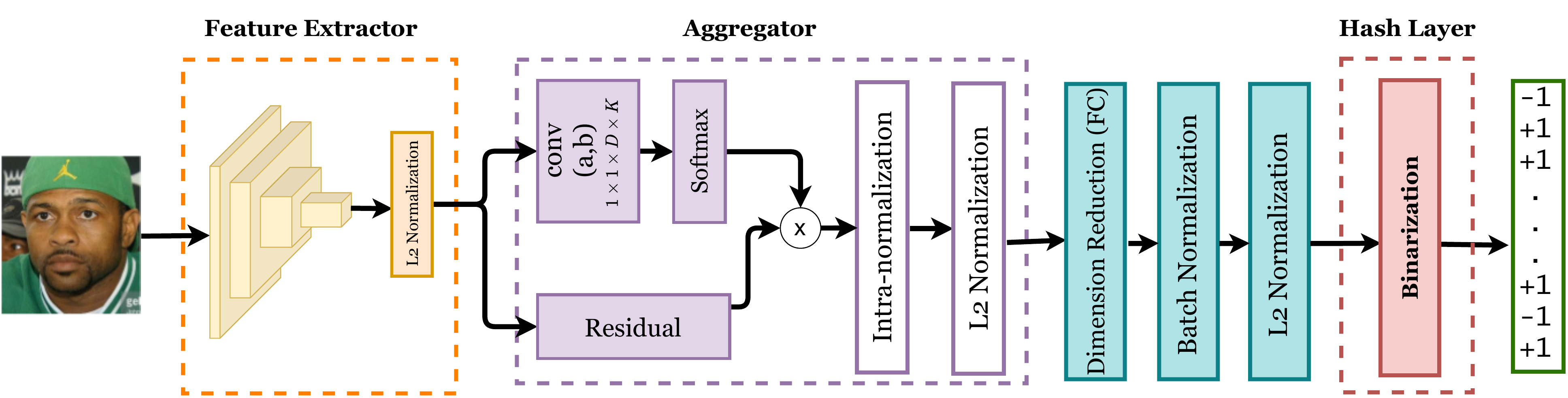}
	\caption{The architecture of AggNet.}
	\label{fig:aggnet}
\end{figure*}

\def \sign {\mathrm{sign}}
We present AggNet, our deep aggregation network. AggNet jointly learns a feature extraction from individual face and an aggregation mechanism merging multiple descriptors into a compact binary code.
Once learned, these functionalities are applied to any group of individuals. The binary code is the data structure representing that group of individuals.
The same network is used to compute a binary hash code for a query face image during verification. 
To compute the membership score, the inner product of the group representation and query is passed to a logistic regression classifier.
Based on this score, access to the system is then granted or denied.
The network architecture is shown in Fig.~\ref{fig:aggnet}, and it consists of:
\begin{itemize}
	\item \textbf{Feature extractor}: a CNN extracting one descriptor from a face image,
	
	\item \textbf{Aggregator}: a learnable pooling layer for aggregating multiple descriptors into a single group representation, and
	
	\item \textbf{Hash layer}: a quantization of the aggregated vector into a binary code.
\end{itemize}
Note that both group and individual feature extraction are derived from AggNet. The following sections discuss each component of the network in detail.

\subsection{Feature Extraction}
Even though the proposed system can be integrated with any feature extraction network, that gives a fixed-length representation, we use an adapted ResNet50 architecture \cite{he2016deep} as the backbone in our work.
We replaced the last fully connected layer of ResNet50 with a convolutional layer with $128$ filters of size $1\times1$.
%
Finally, L2 normalization is applied to each descriptor. 

\subsection{Aggregation}
The feature extraction outputs the individual features $\{\x_{1},\dots ,\x_n\}\subset\mathbb{R}^d$ from a group of $n$ members.
The aggregation is \emph{NetVLAD}~\cite{arandjelovic2016netvlad}, which is a trainable pooling layer based on VLAD~\cite{jegou:inria-00633013}.
This block aggregates multiple features into a $d.K$ vector, where $K$ is the number of centroids in the NetVLAD dictionary.
A fully-connected layer then reduces the dimension back to $d$, the dimension of the feature vectors.
A batch-normalization layer and L2-normalization are also used to produce the aggregated vector.
VLAD and NetVLAD are discussed in more detail in the following.

The Vector of Locally Aggregated Descriptors (VLAD) is a global image descriptor aggregated from local descriptors. The VLAD method starts with quantizing the local descriptors onto a visual vocabulary learned by $K$-means. VLAD takes into account the difference between the local descriptor $\x_i$ and its quantization onto the visual word $\mathbf{c}_k = Q(\x_i)$.
NetVLAD replaces this quantization by a soft assignment.
It first computes a $d \times K$ matrix using the following equation:
\begin{equation}
H_g(j,k)=
\sum_{i=1}^{n}\frac{e^{\textbf{a}_k^T
		\x_i+b_k}}{\sum_{k^\prime=1}^{K}{e^{\textbf{a}_{k^\prime}^T \x_i+b_{k^\prime}}}}
	 {\left(\x_i(j)-\mathbf{c}_k(j)\right)},
\label{netvlad_eq}
\end{equation}
where $\x_i(j)$ and $\textbf{c}_k(j)$ are the $j$-th dimensions of the $i$-th
descriptor and $k$-th cluster center, respectively, and
$\textbf{a}_k$, $b_k$ and $\textbf{c}_k$ for $k \in \{1,\dots,K\}$ are trainable parameters of this block.
The weight of each residual, \ie\ the fraction term, indicates
the soft-assignment of input $\x_i$ to cluster $\textbf{c}_k$.
The final vector $\hh_g\in\mathbb{R}^{d.K}$ is obtained by flattening and L2 normalization.

\subsection{Hashing}
The hashing block takes a real vector, maps it to a rotated space where it coarsely quantizes it onto $\{-1,+1\}^d$. 
One possibility is to first train the mapping network ($\mathcal{F}$), and then apply the $\sign$ function to the learned real-valued representations, as a post-processing.
However, this leads to a sub-optimal solution.
We propose to include the binarization process in the training.
For this, one can add a layer at the end of the network that applies the $\sign$ operation, but its non-differentiability and flatness prevent the learning of $\mathcal{F}$.
%
To tackle this problem, we employ the Greedy hash technique proposed in~\cite{su2018greedy} and explained in Sec.~\ref{subsec:greedy-hashing}.
\subsection{Loss Function}
During the training, we assume that $n$ identities belong to a group and each training batch consists of faces of $m$ individuals.
In the forward pass, the features corresponding to the $n$ individuals will be aggregated into a single binary hash code $\bg$.
Also, for every identity in the batch, a binary hash code, $\bi$, is computed using the same network (as if $n=1$).

The association to the group is then determined for each individual by applying a logistic regression classifier to the scalar product of the two binary codes, \ie:
\begin{equation}
score(\textbf{b}_g,\textbf{b}_i) = \sigma(\lr_1\simm+\lr_2)
\label{score}
\end{equation}
where $\sigma(z)=(1+exp(-z))^{-1}$ is a sigmoid function, and $\lr_1$ and $\lr_2$ are the slope and bias parameters of the logistic regression classifier, respectively. This score should ideally be one
if indiviudal $i$ is a member of group $g$, and zero otherwise.
The loss indicates how much of a difference there is between this ideal score and the actual score.

As usual in machine learning, the choice of loss function plays an important role.  In the following, we define two different loss functions.

\subsubsection{Weighted Cross Entropy Loss}
\label{WCE loss}
For each group, the loss is defined as:
\begin{multline}
\mathcal{L}=\sum_{i=1}^{m} w_{i} [g_i \log(score(\textbf{b}_g,\textbf{b}_i))\\
+(1-g_i)\log(1-score(\textbf{b}_g,\textbf{b}_i))],
\label{loss_bc}
\end{multline}
where $m$ is the size of the training batch, $g_i$ is a binary indicator whether $i$-th individual belongs to the $g$-th group.
Parameter $w_i$ is a weight applied to $i$-th individual  to tackle the data imbalance problem:
There are more negatives than positives in a batch (most $g_i$'s are equal to 0).

\subsubsection{Wilcoxon-Mann-Whitney Loss}
\label{WMW loss}
An ideal loss function directly corresponds to the metric by which we evaluate performance.
For instance, the Area Under the ROC Curve ($AUC$) or the probability of true positive for a fixed false positive rate ($P_{tp}@P_{fp}=u$).
The $AUC$ can be computed using Wilcoxon-Mann-Whitney (WMW) statistic:
\begin{equation}
AUC = \frac{1}{n(m-n)}\sum_{\substack{i,j=1 \\ g_i>g_j}}^{n} I\left(score(\textbf{b}_g,\textbf{b}_i)-score(\textbf{b}_g,\textbf{b}_j)\right),
\label{WMW_equation}
\end{equation}
where $I(z)=\mathds{1}[z>0]$ is the Heavyside unit step function.
However, the $AUC$ is not a differentiable function.
It can be smoothed out and become differentiable, as discussed in \cite{yan2003optimizing}, by approximating Eq.~\eqref{WMW_equation} by:
\begin{multline}
\label{loss_wmw}
AUC\approx \mathcal{L} =  \frac{1}{n(m-n)}\sum_{\substack{i,j=1 \\ g_i>g_j}}^{n} R(\bi,\bj),\\
\mbox{with}\quad
R(\bi,\bj) =  \begin{cases}
(-(S_{ij}-\gamma))^p & \text{if } S_{ij} < \gamma\\
0 & \text{o.w.}
\end{cases}
\end{multline}
and $S_{ij}=score(\textbf{b}_g,\textbf{b}_i)-score(\textbf{b}_g,\textbf{b}_j)$.
The hyperparameters are $\gamma\in(0,1)$ (usually $0.1<\gamma\leq 0.7$) and $p > 1$ (usually 2 or 3).
As a loss function, it penalizes any occurrence where the score of a group member is less than the one of a non-member. 

\subsubsection{Greedy Hashing}
\label{subsec:greedy-hashing}
\def \BB{\mathbf{B}}
Once we have defined the loss function (either~\eqref{loss_bc} or~\eqref{loss_wmw}), learning amounts to find the parameters of the feature extraction and hashing minimizing the loss.
The main difficulty is that the loss is measured from binary vectors which prevents a gradient descent.
A first idea is to solve the problem first on the binary vectors stacked in matrix $\BB = [\bb_1,\ldots,\bb_N]$, \ie\
\begin{equation}
\min_{\BB\in\{-1,1\}^{\bin\times N}} \mathcal{L}(\BB),
\end{equation}
and then to find the extraction and hashing functions that provide this solution.
Yet, that problem is a priori NP-hard.
Paper~\cite{su2018greedy} gives a suboptimal but tractable solution, so-called greedy hashing.
The non-differentiable operator  (in our case the sign function) is taken into account in the forward pass of the hash layer, but not in the backward propagation.
This means that the gradient of any $\bb_j$ is transmitted to $\hh_j$ entirely \cite{su2018greedy}.
This amounts to set $\hh_j^{t+1}$, the output of the NetVLAD layer, and $\bb_j^{t+1}$ its binarization at iteration $t+1$ to:
\begin{eqnarray}
\bb^{t+1}_j&=&\func{sgn}(\hh^{t+1}_j),\\
\hh^{t+1}_j&=&\bb^{t}_j- \alpha \frac{\partial \mathcal{L}}{\partial \mathcal{\bb}_j^t}.
\label{split_greedy}
\end{eqnarray}
The greedy approach of~\cite{su2018greedy} also advices to add a penalty term $\sum_j\|\hh_j-\func{sgn}(\hh_j)\|_3^3$ in the loss function in order to force the components of any $\hh_j$ is be closer to $\{-1,1\}$
whence limiting the impact of the coarse quantization.

\section{Experimental Results}
\label{sec:exp}

\subsection{Datasets}
We used multiple face datasets whose 
images are taken in unconstrained settings, exhibiting significant variation across a variety of variables (e.g., pose, age, gender, facial expression).\\ 
\textbf{VGGFace2} comprises 3.31 million images of 9,131 identities, with on average 360 images for each identity. The images are downloaded from Google Image Search.
A training set of 8,631 identities and a test set, including 500 identities are created from the dataset.\\
\textbf{CelebFaces+} includes 202,599 face images of 10,177 identities (celebrities) collected from the Internet. The dataset consists of a train (including 162,770 images), validation (including 19,962 images) and test set (including 19,962 images).\\
\textbf{LFW} or the `Labelled Faces in the Wild', contains 13,233 face images of 5,749 identities collected from the Internet. We use LFW only for the testing.\\
\textbf{CASIAWebFace} contains about 10K celebrity identities with a total of 500K images collected from the IMDb website. We use CASIAWebFace only for the testing.

%
%
We use images from the training partition of VGGFace2 to train the network, and images from its test partition as the validation set. 
We also removed the overlapping identities from the test datasets to report fair results.

\begin{figure}[h]
	\centering
	\includegraphics[width=0.8\linewidth]{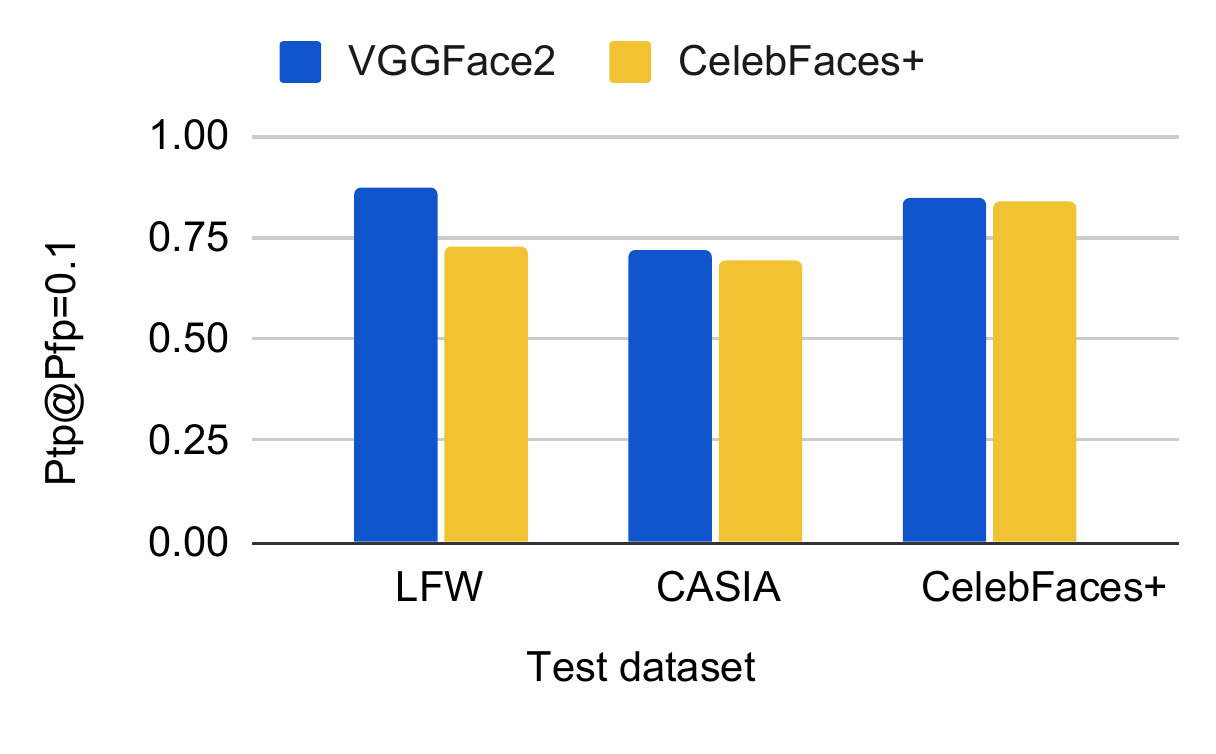}
	\caption{Impact of the training set.}
	\label{fig:CelebVsVGG}	
\end{figure}
\subsection{Implementation details}
\begin{table*}[!h]
	\centering
	\begin{tabular}{c c c c||c c c c c} 
		\multicolumn{4}{c||}{Ablation} & \multirow{2}{*}{$Accuracy$} & \multicolumn{3}{c}{$\ptp@\pfp$} & \multirow{2}{*}{$AUC$} \\
		& training & loss & pooling  & & 1\% & 5\% & 10\% & \\
		\hline
		EoA \cite{Gheisari2019icassp} & \scriptsize \xmark & - & \scriptsize Sum & 0.74 & 0.15 & 0.33 & 0.45 & 0.80 \\
		Sum-WMW & \scriptsize  \cmark & \scriptsize WMW & \scriptsize Sum & 0.90 & 0.36 & 0.65 & 0.80 & 0.93\\ 
		GeM-WMW & \scriptsize \cmark & \scriptsize WMW & \scriptsize GeM & 0.91 & 0.36 & 0.66 & 0.80 & 0.93\\
		AggNet-WCE & \scriptsize \cmark & \scriptsize WCE & \scriptsize NetVLAD & \textbf{0.92} & 0.34 & 0.68 & 0.83 & 0.93 \\ 
		AggNet-WMW & \scriptsize \cmark & \scriptsize WMW & \scriptsize NetVLAD & \textbf{0.92} & \textbf{0.46} & \textbf{0.75} & \textbf{0.87} & \textbf{0.95}\\ 
	\end{tabular}
	\caption{Ablation study}
	\label{table:ablation}
\end{table*}
\begin{figure*}[h]
	\centering	
	\begin{subfigure}{0.32\linewidth}
		\includegraphics[width=\textwidth] {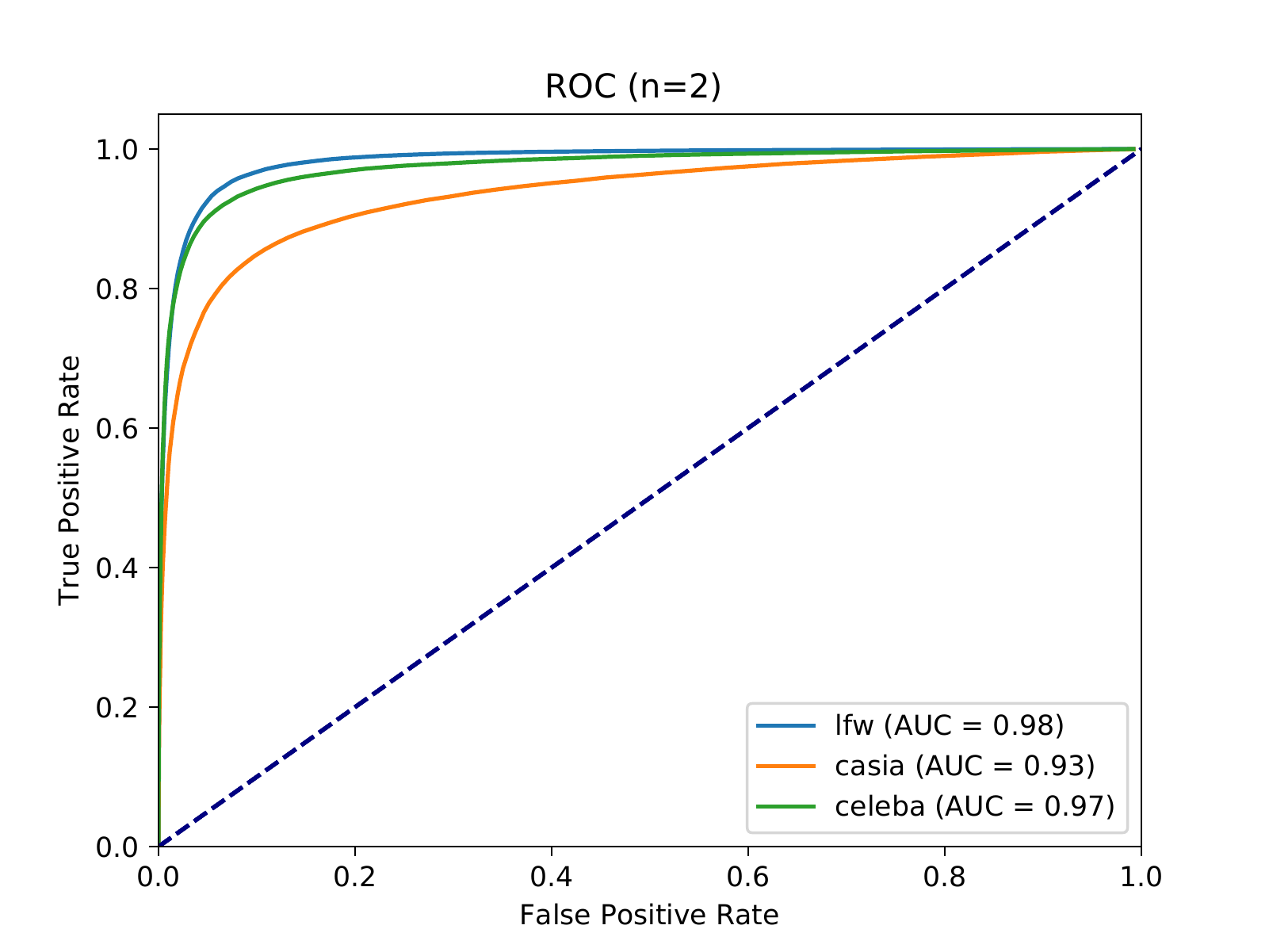}
	\end{subfigure}
	\hfill
	\begin{subfigure}{0.32\linewidth}
		\includegraphics[width=\textwidth] {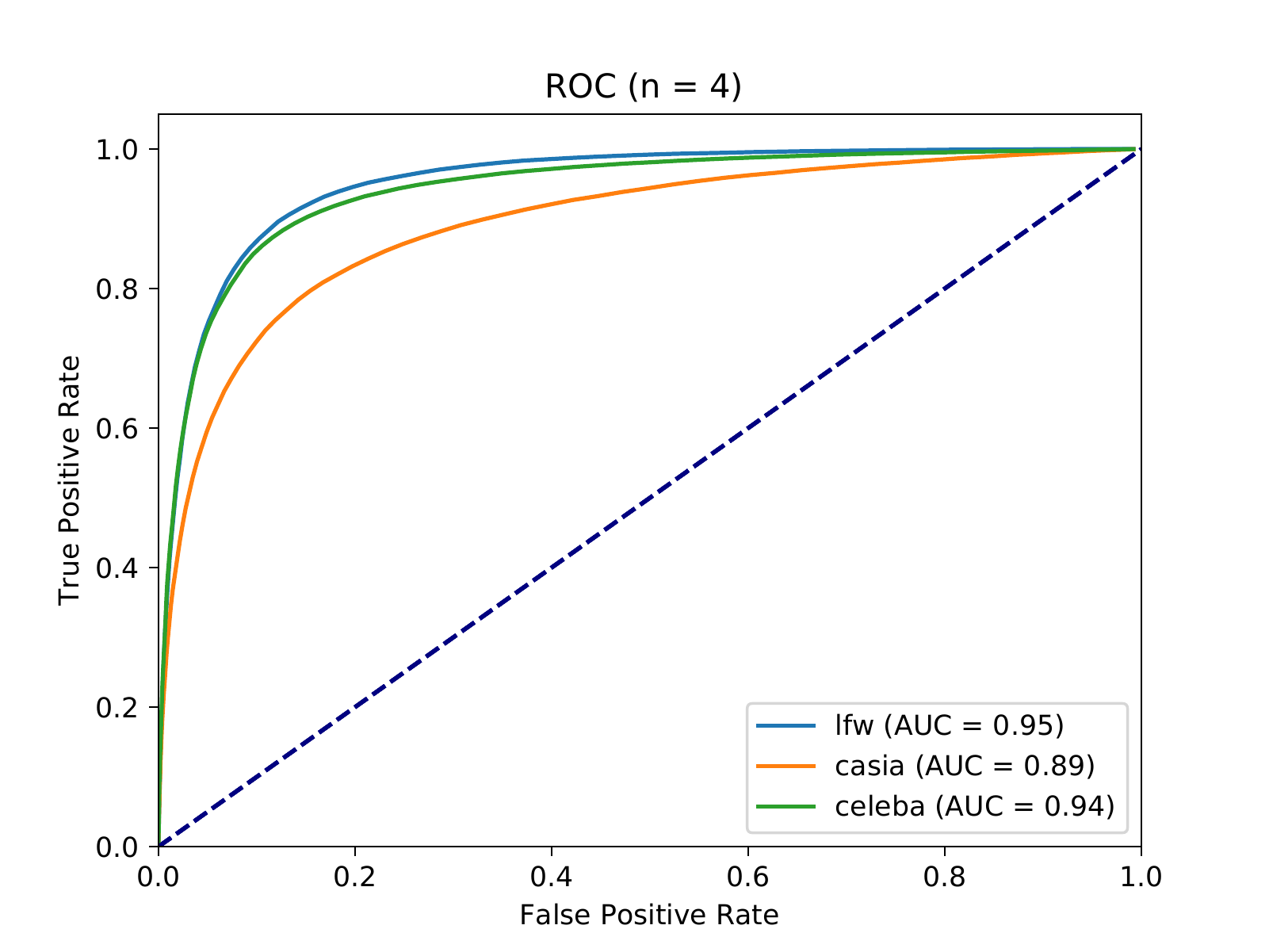}
	\end{subfigure}
	\hfill
	\begin{subfigure}{0.32\linewidth}
		\includegraphics[width=\textwidth] {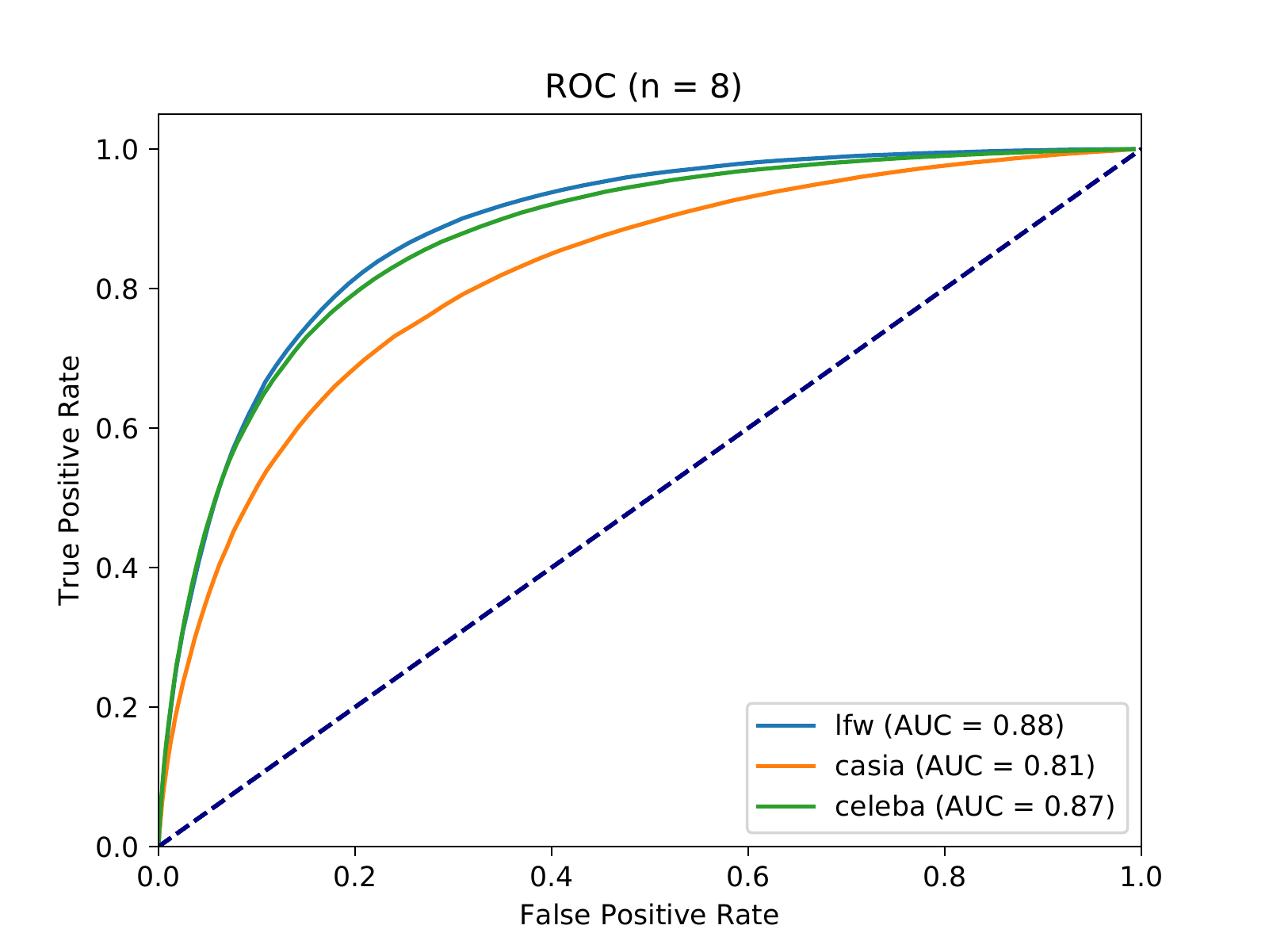}
	\end{subfigure}
	\caption{ROC curves for AggNet with different group sizes}
	\label{fig:ROC_m}
\end{figure*}
Here are the details about the training and testing phases.\\
\textbf{Initialization.}
The face feature extraction is pretrained using softmax loss for face classification problem on VGGFace2. 
Initialization of NetVLAD is done using the $K$-means algorithm with $K = 8$ clusters. 
At both training and test time, the input image is first resized to $256\times256$ and central cropped to $224\times224$ before feeding the network.\\
\textbf{Training.}
Training is done on the representations of groups constructed by randomly sampling identities.
For each identity in the group, two faces are randomly sampled.
One is enrolled and contributes to producing the group representation.
The other is used as a query, and its scalar product is computed against all the group representations in the same batch.
We use Stochastic Gradient Descent to train the network, with weight decay $0.001$, momentum $0.9$, and learning rate of $0.001$.
In the following epochs, if the validation loss is not improving, the learning rate is reduced by a factor of $0.1$. The batch size is 64.\\
\textbf{Group Membership Verification.} 
At enrollment, the group representations are produced for each group.
Then, at verification time, the face descriptor of the query is compared with group representation corresponding to the claimed group by using Eq.~\eqref{score}.
\subsection{Results}
We report the AUC, or `Area under the ROC Curve' and  $\ptp$ $@1\%$, $\ptp@5\%$ and $\ptp@10\%$ where $\ptp @ (\pfp=u)$ is the `Probability of True Positive' when `Probability of False Positive' is equal to $u$. It means that the score are compared to a threshold set to a value ensuring such a probability of false positive.
As for the accuracy, the queries with a score above the threshold $0.5$ are deemed group member.
The accuracy is then measured as the ratio of good decisions (true positives and negatives).
%
In the following experiments the group size $n$ is equal to 4.

\subsubsection{Ablation study}
The ablation study on LFW dataset assesses the impact of the model components on the performances of AggNet.
Table~\ref{table:ablation} shows the results.  
As a \emph{baseline}, we take the  Embedding of Aggregation (EoA) scheme of~\cite{Gheisari2019icassp}.
In this model, the feature extractor is pretrained, a sum pooling is used for the aggregation and finally, the binarization is done using a sign function.
The baseline does not include any trainable blocks except the feature extractor.

The Sum-WMW and GeM-WMW variants have the same training procedure as the AggNet-WMW.
The only difference is that the NetVLAD pooling mechanism is replaced with a Sum and GeM pooling, respectively. GeM (Generalized mean pooling) \cite{radenovic2018fine} is a variation of Sum pooling that adjusts the contribution of each feature vector through a learnable parameter.

The \textit{loss} column in the table indicates the loss function used for training the model: Wilcoxon-Mann-Whitney (WMW~\eqref{loss_wmw}) or Weighted Cross Entropy (WCE~\eqref{loss_bc}).


We can see that NetVLAD achieves overall better results and GeM with the learnable parameter has no advantage over simple sum pooling.
It can be attributed to the fact that GeM is a very simple aggregation technique and cannot perform as well as NetVLAD, a considerably more sophisticated method.
Also, the local feature extraction network is not able to produce informative features, therefore once aggregated using a simple GeM layer, it does not produce a discriminative group representation.
Further, minimization of WMW results in the same accuracy level as minimization of WCE, but it significantly improves AUC and Probability of True Positive scores. This can be explained by the fact that the WML loss directly optimizes the AUC metric while the minimization of cross-entropy does not necessarily improve the AUC. As a result, in applications where the true positive rate is the metric of interest, it is suggested to directly optimize the AUC of the classifier.


\subsubsection{Impact of training data}
The AggNet model is trained on both VGGFace2 and CelebFaces+ datasets.
This way, the impact of the training data can be evaluated at test time (on different datasets).
Both models use Res-Net50 and the same training procedure ; the only difference is the data used for training. The $\ptp@\pfp=0.1$ results are compared in Fig. \ref{fig:CelebVsVGG}.
Training on VGGFace2 always achieves higher  performances, especially when tested on LFW dataset. A larger dataset allows the network to generalize more effectively to new data.

\subsubsection{Impact of group size}
Fig.~\ref{fig:ROC_m} shows the ROC curves when $n=2$, $4$, or $8$. For each of those, the model is trained separately.
Clearly, packing more faces into groups inevitably results in information loss, and thus lowers performance.

\subsubsection{Impact of binarization (vs. real valued output)}
\begin{figure}[h]
	\centering	
	\begin{subfigure}{0.45\linewidth}
		\includegraphics[width=\textwidth] {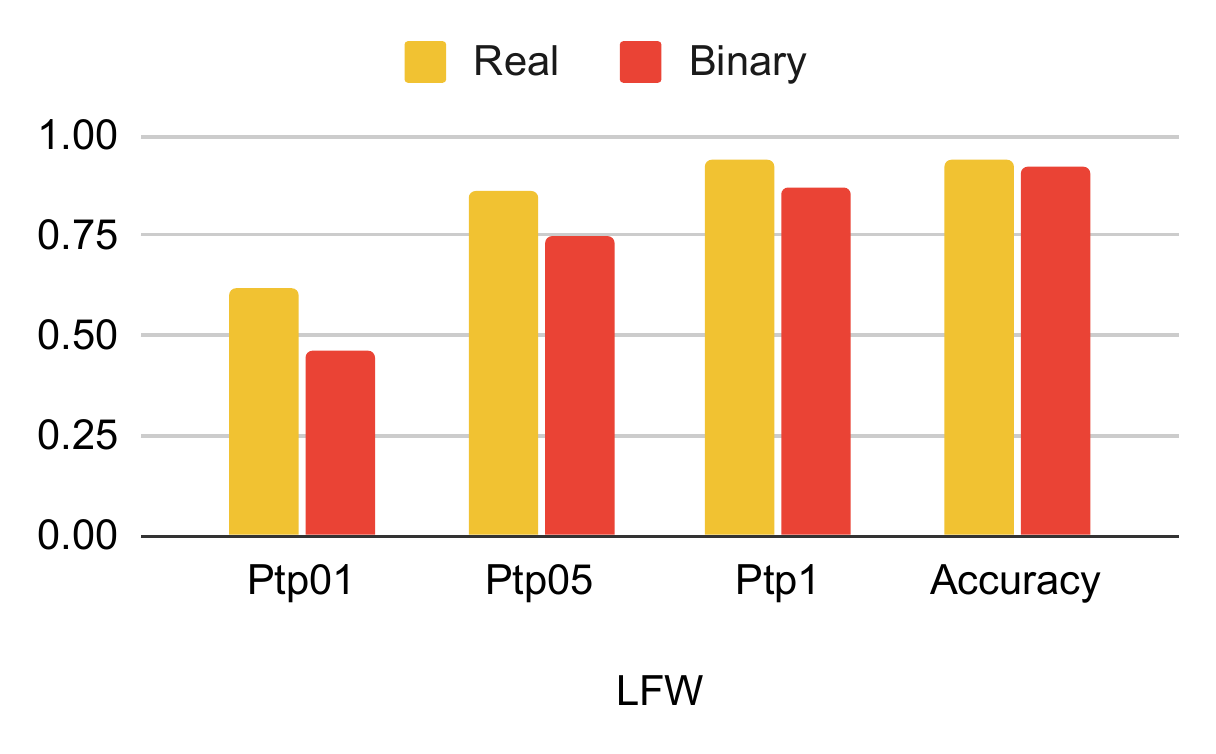}
	\end{subfigure}
	\begin{subfigure}{0.45\linewidth}
		\includegraphics[width=\textwidth]
		{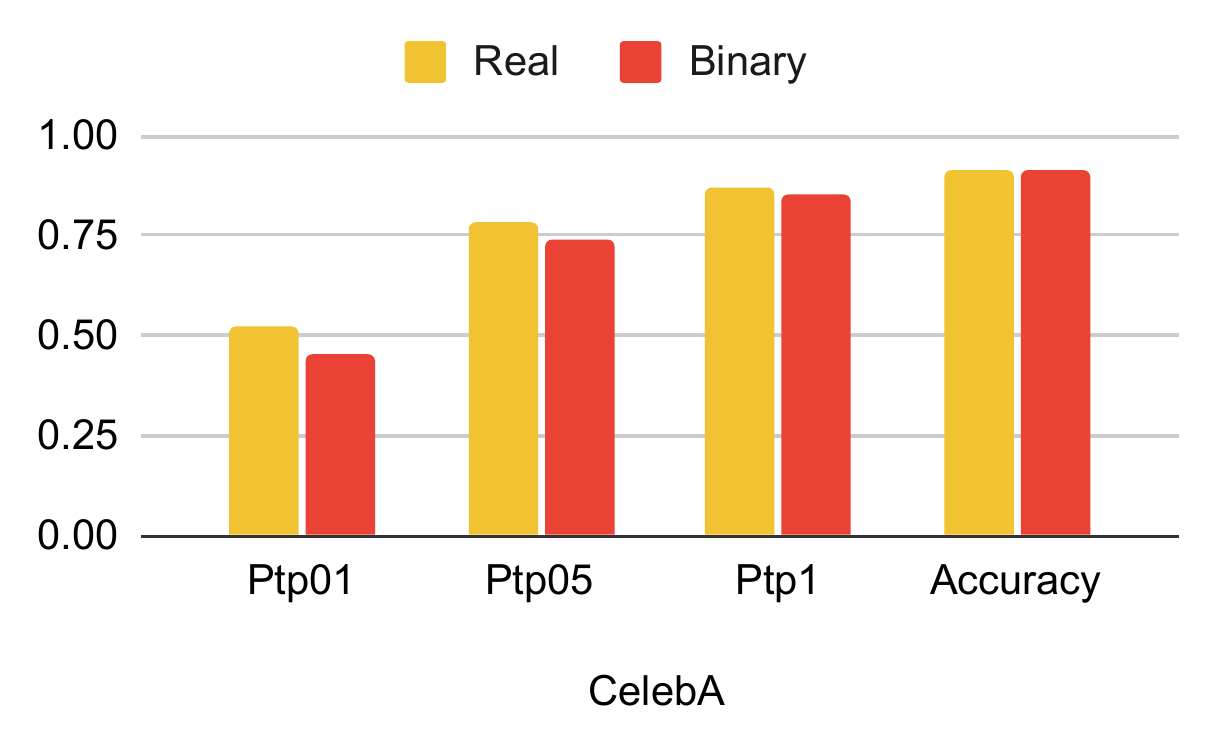}
	\end{subfigure}
	\caption{Comparison of binary vs. real-valued representations}
	\label{fig:BinaryVsReal}
\end{figure}
\begin{figure}[h]
	\centering
	\includegraphics[width=0.95\linewidth]{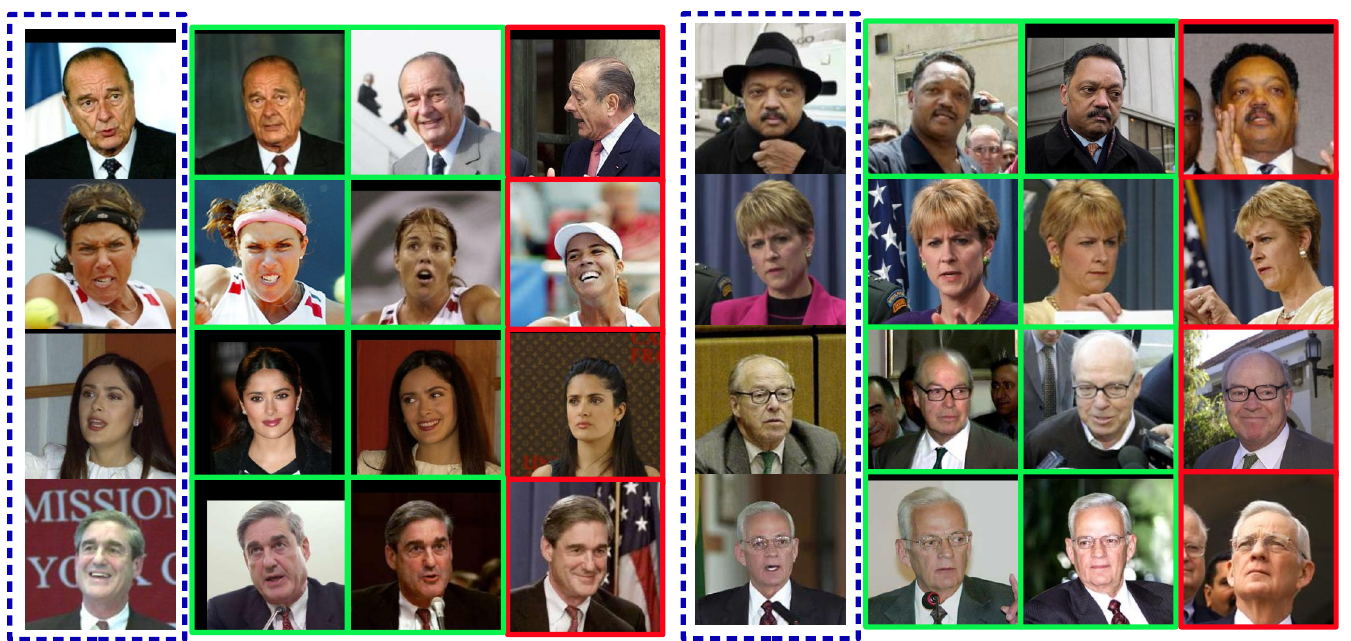}
	\caption{Some examples of group membership verification on LFW with $n=4$. Samples in blue frames are enrolled, the samples in green/red frames are successful/failed queries, respectively.}
	\label{fig:lfw_examples}
\end{figure}
We now train AggNet by removing the binarization layer, so that we obtain real-valued descriptors instead of binary ones.
Fig.~\ref{fig:BinaryVsReal} compares the two variants on LFW and CelebFaces+ datasets.
Binarization degrades the performances due to the information loss and the suboptimal learning strategy (Sect.~\ref{subsec:greedy-hashing}).
Yet, the drop is surprisingly small.
Binary descriptors are desirable for two reasons: compactness and compliance with lightweight homomorphic encryption.
In other words, computing the score of an encrypted query is feasible, while the group representation remain in the clear.
Aggregation and binarization protect the enrolled templates because they prevent their reconstruction.


\section{Conclusion and Future Work}
\label{sec:conclusion}
In this paper, we address the problem of group membership verification. The proposed architecture, AggNet, jointly learns face descriptors and the aggregation mechanism in an end-to-end manner. 
We tested on multiple large scale face datasets.
Our study shows that AggNet encodes distinctive information with the less amount of information loss and provides better generalization capability when compared to other baselines. 
We also used the Wilcoxon-Mann-Whitney loss which directly maximizes the AUC metric, 
and demonstrated that the verification performance is boosted by using this loss function.
Moreover, the proposed architecture can be applied to other similarity search problems such as image retrieval, where $N$ data vectors are packed into $M<<N$ memory vectors. 
The system yields substantial gains in space (memory footprint) and time (query runtime).
In the future, we will investigate the requirements for packing higher numbers of members to each group.

\bibliographystyle{IEEEtran}
\bibliography{egbib}
%
%

\end{document}